\documentclass[letterpaper, 10 pt, conference]{ieeeconf}  

\IEEEoverridecommandlockouts                              

\usepackage{ifpdf}
\usepackage{graphicx}
\ifpdf
\usepackage{amsmath}
\usepackage{autobreak}
\usepackage{hhline}
\usepackage{multirow}
\usepackage{hyperref}
\usepackage{graphicx}

\usepackage{algorithmic}
\usepackage{algorithm}  
\usepackage{dblfloatfix}
\usepackage{blindtext}

\usepackage{amsmath}

\graphicspath{{Figs/}{Figs/PDF/}}
\else
\graphicspath{{Figs/}}
\fi
\usepackage{hhline}
\usepackage{color}
\usepackage{calc}
\let\labelindent\relax
\usepackage{enumitem}
\usepackage{bm,amsmath}

\usepackage{nomencl}
\usepackage{multirow}

\usepackage{url}
\usepackage{nomencl}
\usepackage{multirow}
\usepackage{multicol}
\usepackage{multirow}

\usepackage{bbding}
\usepackage{booktabs}
\usepackage{makecell}
\hypersetup{
	colorlinks = true,
	linkcolor = blue,
	citecolor = blue
}
\usepackage{subcaption}

\usepackage{xcolor}

\usepackage{caption}
\usepackage[flushleft]{threeparttable}
\usepackage{cite}
\usepackage{times}
\usepackage{epsfig}
\usepackage{graphicx}
\usepackage{amsmath}
\usepackage{amssymb}

\usepackage[maxfloats=256]{morefloats}
\title{\LARGE \bf
Learn from Incomplete Tactile Data: Tactile Representation Learning with Masked Autoencoders
}

\author{Guanqun Cao$^{1}$, Jiaqi Jiang$^{ 2}$, Danushka Bollegala$^{1}$, and Shan Luo$^{2}$
\thanks{This work was funded in part by the EPSRC project ``ViTac: Visual-Tactile Synergy for Handling Flexible Materials'' (EP/T033517/2).}
\thanks{$^{1}$G. Cao and D. Bollegala are with the Department of Computer Science, University of Liverpool, Liverpool L69 3BX, United Kingdom. Emails:~\{\tt\small g.cao, danushka\}@liverpool.ac.uk.}
\thanks{$^{2}$J. Jiang and S. Luo are with the Department of Engineering, King's College London, London WC2R 2LS, United Kingdom. E-mails: {\tt\small \{jiaqi.1.jiang, shan.luo\}@kcl.ac.uk}.}
}

\begin{document}

\maketitle
\thispagestyle{empty}
\pagestyle{empty}

\begin{abstract}

The missing signal caused by the objects being occluded or an unstable sensor is a common challenge during data collection.
Such missing signals will adversely affect the results obtained from the data, and this issue is observed more frequently in robotic tactile perception.
In tactile perception, due to the limited working space and the dynamic environment, the contact between the tactile sensor and the object is frequently insufficient and unstable, which causes the partial loss of signals, thus leading to incomplete tactile data.
The tactile data will therefore contain fewer tactile cues with low information density.
In this paper, we propose a tactile representation learning method, named \textit{TacMAE}, based on Masked Autoencoder to address the problem of incomplete tactile data in tactile perception.
In our framework, a portion of the tactile image is masked out to simulate the missing contact region. 
By reconstructing the missing signals in the tactile image, the trained model can achieve a high-level understanding of surface geometry and tactile properties from limited tactile cues.
The experimental results of tactile texture recognition show that our proposed TacMAE can achieve a high recognition accuracy of 71.4\% in the zero-shot transfer and 85.8\% after fine-tuning, which are 15.2\% and 8.2\% higher than the results without using masked modeling. 
The extensive experiments on YCB objects demonstrate the knowledge transferability of our proposed method and the potential to improve efficiency in tactile exploration.

\end{abstract}

\section{INTRODUCTION}

The tactile properties of an object's surface are important information for robots to gain an understanding of the physical environment.
The surface tactile properties, such as textures, stiffness, softness, etc., are embedded into the tactile data received by the physical interaction between the tactile sensor and the target objects, which enable robots to facilitate manipulation tasks and interact with their surrounding effectively~\cite{jiang2022shall}.

During the tactile data collection process, the tactile sensor is expected to have adequate contact with the target objects, leading to the activation of  larger perception fields and distinct tactile signals to be recorded.
However, due to the dynamic environment and limited working space for robotics, it often happens that the tactile sensor fails to make contact or only makes partial contact with the target object, especially for soft objects like clothes. 
In such a case, the recorded incomplete tactile data contains fewer tactile cues than the data from adequate contact because of a smaller perception field being stimulated.
Moreover, some tactile properties, such as compressibility and stiffness, which need to be obtained by squeezing the target, would be absent from partial contact.

In tactile perception, the tactile image collected from the optical tactile sensor is one of the most popular tactile data types~\cite{yuan2017gelsight,ward2018tactip}.
A typical example is the tactile image from the GelSight sensor,  which has embedded tactile textures, height information, and friction information with high resolution.
Since tactile images share the same data format as RGB images, they have been processed by using advanced techniques from the field of computer vision\cite{yuan2018active,cao2020spatio,polic2019convolutional}.
However, most methods are limited to treating adequate and partial contact events equally.
We argue that the current methods ignore the effect of the incomplete tactile data and there is no specific optimisation or representation method to address this issue.

\begin{figure}[t]
	\centering
	\includegraphics[width=0.8\columnwidth]{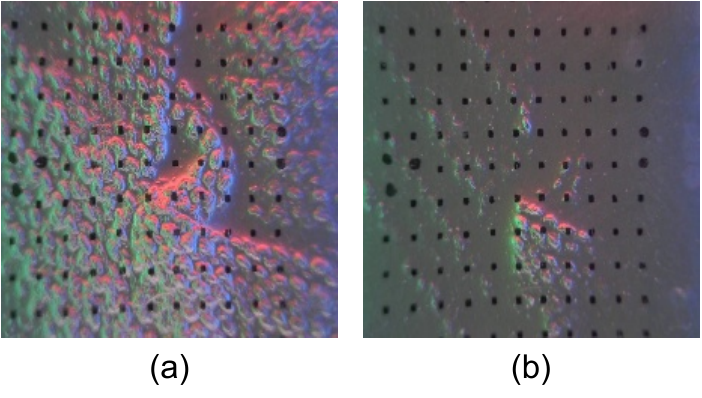}
	\caption{\textbf{\textit{Tactile images from the GelSight sensor.}} (a) A tactile image from adequate contact; (b) An incomplete tactile image from partial contact. Compared with adequate contact, there are much fewer tactile cues embedded in partial contact. The black markers on the sensor are used to improve the spatial acuity in tactile perception.  }
	\label{fig:first}
\end{figure}

As shown in Fig.~\ref{fig:first}, the major difference between tactile data from adequate contact versus partial contact comes from the information density, where some contact regions are missing in the partial contact.
However, humans are able to identify an object by touching a small portion of the object's surface area~\cite{klatzky1989haptic}.
It indicates that the tactile signals have a high degree of redundancy on the object's surface and the patterns of the object's surface tend to have repetitive parts with identical properties.
Consequently, it is possible to reconstruct a missing signal in tactile data from neighbouring signals in the spatial space using semantic information of of the existing tactile cues.

In this paper, we propose a tactile representation method based on the Masked Autoencoder~\cite{he2022masked}, named \textit{TacMAE}, to simulate the contact area's absence of incomplete tactile data caused by partial contact. 
Motivated by a recent work~\cite{xie2022masked}, which uses the amount of the remaining voxels to represent the degree of occlusion in 3D point cloud, we employ a similar approach that uses contact area in tactile images as the criterion to determine adequate contact and partial contact.
In the training stage, a portion of the input patches of tactile images is masked out using the data collected from adequate contact, and the missing tactile signals are then reconstructed.
Moreover, a supervised classification head is designed, which allows us to learn additional information from corresponding labels.
After training, the encoder can be applied to the incomplete tactile images to learn the latent representations with low information density effectively.
In the tactile texture recognition experiments, we observe a significant improvement in recognising the objects with partial contact, in both zero-shot transfer and fine-tuning settings.
Moreover, our proposed method can be seen as a sensor-agnostic representation method by converting the signals into the image format from any other tactile sensor.

The contributions of this paper are summarised as follows:
\begin{enumerate}

    \item We propose a tactile representation method \textit{TacMAE} to address the problem of incomplete tactile data in robotic perception, which is the first of its kind.
  
    \item  We use a masked autoencoder with a high masking ratio to simulate the absence of contact area in partial contact. By reconstructing the missing signals using observed information, the model can effectively learn from incomplete tactile data.
    To the best of the authors' knowledge, no previous studies have investigated mask modeling in tactile representation learning.
    \item The experimental results demonstrate that our method can significantly improve tactile texture recognition performance by learning more robust tactile representations and the experiments of transfer learning indicate our method can be used to improve the efficiency of tactile exploration.
\end{enumerate}

\section{Related works}
In this section, we will first review the works on tactile perception with tactile images, followed by a discussion of masked modeling in representation learning. 
\subsection{Tactile perception with tactile images}
Tactile sensing has been wildly used in robotic exploration with different kinds of sensing mechanisms, such as strain gauges~\cite{jamali2010material}, capacitive sensors~\cite{taunyazov2019towards}, and microphones~\cite{luo2017knock}.
The motion of the sensor provides tactile sensory information of the contacting surface, such as friction information and textures for tactile perception.
Compared with other tactile sensors, optical tactile sensors use high-resolution cameras and record more detailed tactile information in tactile images.
Recently, camera-based optical tactile sensor, such as the GelSight sensor, has gained popularity in tactile perception tasks.
In~\cite{yuan2018active}, the GelSight is applied to enable the robot to recognise clothes and their corresponding properties autonomously.
In~\cite{cao2020spatio}, a spatio-temporal attention model is proposed to process tactile images from the GelSight sensor, which is capable to highlight salient tactile features in both spatial and temporal dimensions for texture recognition.
In \cite{luo2018vitac}, tactile images are fused with visual images to learn the sharing features between vision and tactile sensing for cloth texture recognition.
However, the previous methods treat each contact event equally, ignoring the effect of incomplete tactile data from partial contact for tactile perception.


\subsection{Masking modelling in representation learning}
In representation learning, masking has been used as a way of masking some of the input and training the model to predict the masked input so as to improve the model's capability when some input is absent. Two popular paradigms are Masked Language Modeling (MLM) and Masked Image Modeling (MIM).

MLM has become a successful paradigm in the field of NLP, such as BERT~\cite{devlin2018bert}, RoBERTa~\cite{liu2019roberta}, and GPT models\cite{radford2018improving, brown2020language}, etc.
Models are trained to predict the value of masked tokens of input sentences in order to understand the context of the sentences.
Due to their superior performance, MLMs have been applied for a variety of downstream tasks, including machine translation~\cite{yang2020towards}; speech recognition~\cite{shin2019effective}, question answering~\cite{yang2019end}, and sentiment analysis~\cite{hoang2019aspect}.

MIM also exhibits great potential in the field of computer vision. 
Similar to the mechanism in MLM, the representation is learnt by predicting the missing information from the remaining cues.
In~\cite{he2022masked}, patches of raw pixels are masked by a high masking ratio which are then reconstructed by using visible patches. 
In~\cite{baevski2022data2vec}, the latent representation is predicted based on the view of masked input. 
Additionally, MIM has benefited various visual applications.
In~\cite{bielskimove}, a segmentation method is developed based on top of the MAE by leveraging the synthesized image with shifted objects.
In~\cite{zhangpoint}, a multi-scale MAE framework is proposed to learn the 3D point clouds for shape classification and object detection.   
In~\cite{wang2022facemae}, face privacy is considered and reserved by masking the face images when training the face recognition model.

In representation learning, the concept of the dropout method~\cite{srivastava2014dropout} is similar to the masking modeling, i.e., dropping a portion of the elements to improve the generalisation ability.
However, these two techniques have a difference in their approaches to drop elements: the masking modeling drops a portion of the input and reconstructs missing content while the dropout method discards random neurons in a layer.
It results in different effects on the capabilities of the model that the masking modeling enhances the representation learning of data and the dropout prevents the overfitting from the training data.
To the best of the authors' knowledge, there are no prior works that apply mask modeling in tactile representation learning.
In this work, we develop a tactile representation method based on masked modeling to solve the problem caused by partial contact in tactile perception, for the first time.

\begin{figure*}[t]
	\centering
	\includegraphics[width=1.8\columnwidth]{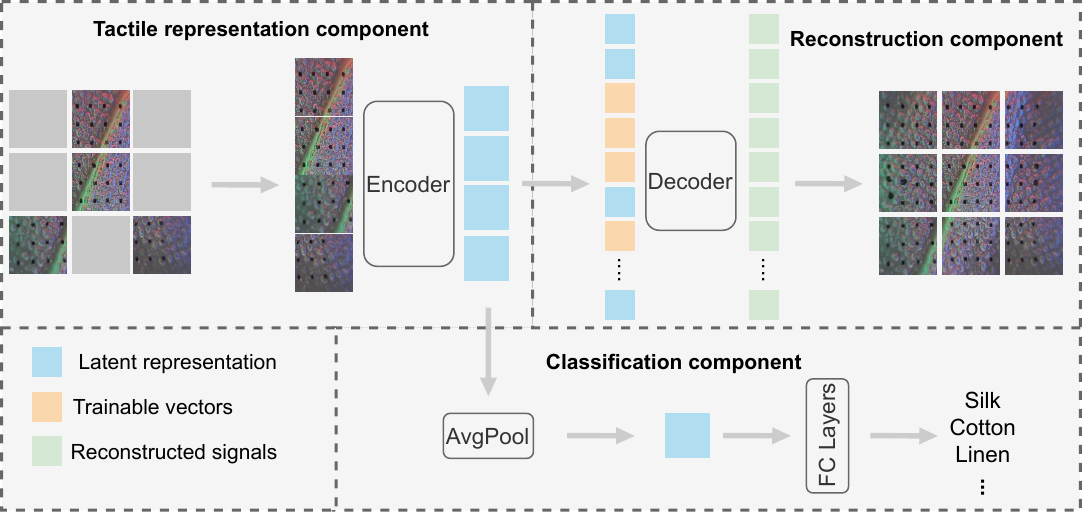}
	\caption{\textbf{Illustration of the proposed TacMAE.} During training, a portion of tactile image patches are randomly masked out.  The encoder encodes the unmasked patches to obtain the latent tactile representation. The decoder is used to reconstruct the masked tactile patches and the classification head is implemented to identify the category of the input by using the latent representation. Trainable vectors are applied to indicate the existence of masked patches for reconstruction.
	} 
	\label{fig:framework}

\end{figure*}
\section{Methodologies}

Our proposed TacMAE masks a portion of signals to simulate the contact area missing in partial contacts and reconstructs the missing signals from limited tactile cues.
As shown in Fig.~\ref{fig:framework}, our framework mainly includes three parts: 1) an encoder $E$ that encodes the unmasked patches of tactile images to obtain latent representation; 2) a decoder $D$ to reconstruct the missing patches of tactile images as well as 3) a classification head to classify the unmasked patches of input tactile image.

\textbf{TacMAE encoder.} Let $\mathbf{x} \in \mathbb{R}^{H \times W \times C}$ denotes the input tactile image obtained from the adequate contact, where $H,W,C$ represent the height, width and channel of the tactile image, respectively.
First, the input tactile image is reshaped into $N$ patches (tokens) $\mathbf{x}_p \in \mathbb{R}^{N \times\left(P^2 \cdot C\right)}$, as token embeddings, where $(P,P)$ represent the shape of each patch and $N=H W / P^2$. 
Then, a portion (e.g., 70\%) of patches are masked out and the remaining unmasked patches are fed into the encoder to obtain the latent representation.
Specifically, we use the structure of Vision Transformer (ViT)~\cite{dosovitskiy2020image} as the encoder in our proposed TacMAE.

\textbf{TacMAE decoder.} A decoder is applied to reconstruct missing patches of tactile images from the unmasked patches, which enables the model to learn from limited tactile cues.
Specifically, our TacMAE receives two components as the input: latent features from the encoder and trainable vectors that illustrate the existence of the missing patches for reconstruction~\cite{he2022masked}.

Concretely, the decoder network is also implemented based on the ViT block.
The decoder reconstructs the tactile image and the mean square error (MSE) is calculated between the original tactile patches and reconstructed tactile patches. 
The reconstruction loss can be represented as:
\begin{equation}
\mathcal{L}_{rec}=\frac{1}{|\Omega|} \sum_{p \in \Omega}|x_p-\hat{x}_p|^2,
\end{equation}
where $p$ denotes the index of masked patches, $\Omega$ represents the number masked patches, $x_p$ represents original value of the masked patches, and $\hat{x}_p$ represents the reconstructed patches.

\textbf{TacMAE classification head.} Apart from the reconstruction by the decoder that focuses on the correlation of the surface geometry in the tactile image, we also apply a classification head to make the model learn additional information from the corresponding labels.
Unlike traditional supervised learning that utilises all patches of tactile images, our approach employs only the latent features of unmasked patches during the training phase. This is due to the fact that the surface patterns of objects often contain repetitive elements with redundancies.
For example, when humans use tactile sensing to recognise the contacting objects, the perceptive area of the tactile area is usually much smaller than the object surfaces~\cite{klatzky1989haptic}, which means that tactile images collected from neighbouring areas of an object surface could be of similar textures.
Consequently, it is possible to train the model to learn from partial tactile information effectively.
Specifically, a global pooling function is performed on the latent features first to obtain global representation.  
Then, two fully connected layers are used in the classification head, and each layer is followed by a ReLU activation function.
Consequently, the cross-entropy loss is calculated between the predicted labels and human-annotated labels in supervised training.
The cross-entropy classification loss is given as:

\begin{equation}
\mathcal{L}_{ce}=-\sum_{i=1}^K y_i \log \hat{y}_i,
\end{equation}
where $\hat{y}$ represents the predicted softmax probability, $y$ represents one hot vector of correct category, and $K$ is the number of samples in our dataset.

By minimising both the classification loss and reconstruction loss, a robust tactile representation is learnt from limited tactile cues during the training stage. 
The overall objective can is expressed as:
\begin{equation}
\mathcal{L}=\lambda_{rec}\mathcal{L}_{rec}+\lambda_{ce}\mathcal{L}_{ce},
\end{equation}
where $\lambda_{rec}$ and $\lambda_{ce}$ are set to 1 and 0.01, respectively,  after a grid search.

\textbf{Implementation on downstream tasks.} After training, fine-tuning can be performed based on different downstream tasks, such as recognition and detection. 
It should be noted that only the encoder needs to be retained as a backbone to obtain the tactile representation while other structures will be removed.
Moreover, different from the aforementioned training which uses a subset of divided patches, uncorrupted tactile images are applied in the downstream tasks.

\section{Data preparation}
In tactile perception, one distinct example with a wide range of surface properties is fabric or clothing. 
In this paper, we use the dataset from~\cite{yuan2017connecting}. 
This dataset contains 118 fabrics with a size of 1m×1m which display various properties, such as textures, colors, density, and stiffness.
The dataset contains visual, tactile, and semantic data, of which only tactile data is used in our study.

During the tactile data collection, an optical tactile sensor, a GelSight sensor, is used to collect a sequence of tactile images by pressing against the fabrics when they are placed on a hard plane.
The fabrics are placed with three different appearances for data collection: laying the fabric flat (flat data), laying it with one fold (fold data), and laying it randomly (random data).
Approximately 10 flat data samples, 15 fold data samples, and 15 random data samples are collected for each fabric.

\begin{figure}[t]
	\centering
	\includegraphics[width=1\columnwidth]{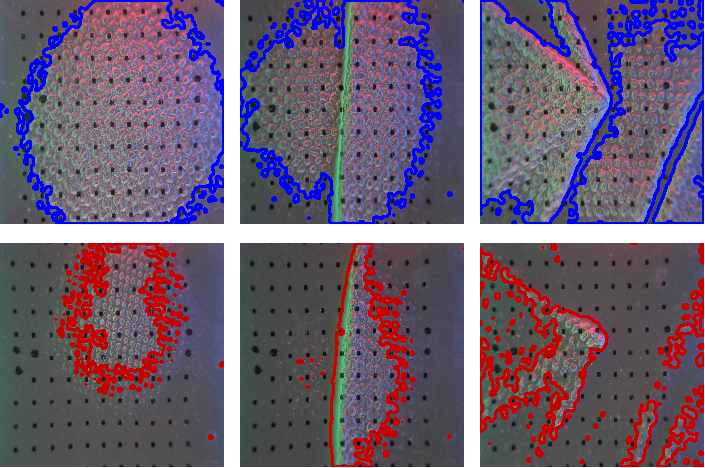}
	\caption{\textbf{\textit{Tactile samples from dataset.}} Tactile images represent adequate contact and partial contact are shown in the top row and the bottom row, respectively. The blue contours and red contours correspond to the contact areas for these two types of interactions. The three columns display the tactile images when the fabric with different stats: flat, folded and laid randomly.  }
	\label{fig:samples}
\end{figure}

To simulate the incomplete tactile images from partial contact, we randomly select three tactile images with contact areas ranging from 10\% to 40\% of the perception field from each contact event.
The tactile images with contact areas over 50\% of the perception field are used as the data from adequate contact.
Some samples of adequate contact and partial contact are shown in Fig~\ref{fig:samples}.
Specifically, we determine the contact area by using the OpenCV \textit{findContours} function.
Accordingly, there are 14,961 tactile images that represent adequate contact and 14,823 tactile images that indicate incomplete tactile data, respectively.
Both datasets are divided into a ratio of 7:2:1 for training, validation and test, respectively.
In this study, we use the contact area as the criterion to distinguish between adequate contact and partial contact. However, there are several alternatives that can be considered, such as contact force or entropy~\cite{tsai2008information} of the tactile image. 
We plan to explore this open question in our future work.

\section{experiments and analysis}
\subsection{Tactile Representation Learning }

\begin{figure}[t]
	\centering
	\includegraphics[trim=40 0 25 25, clip, width=1\columnwidth]{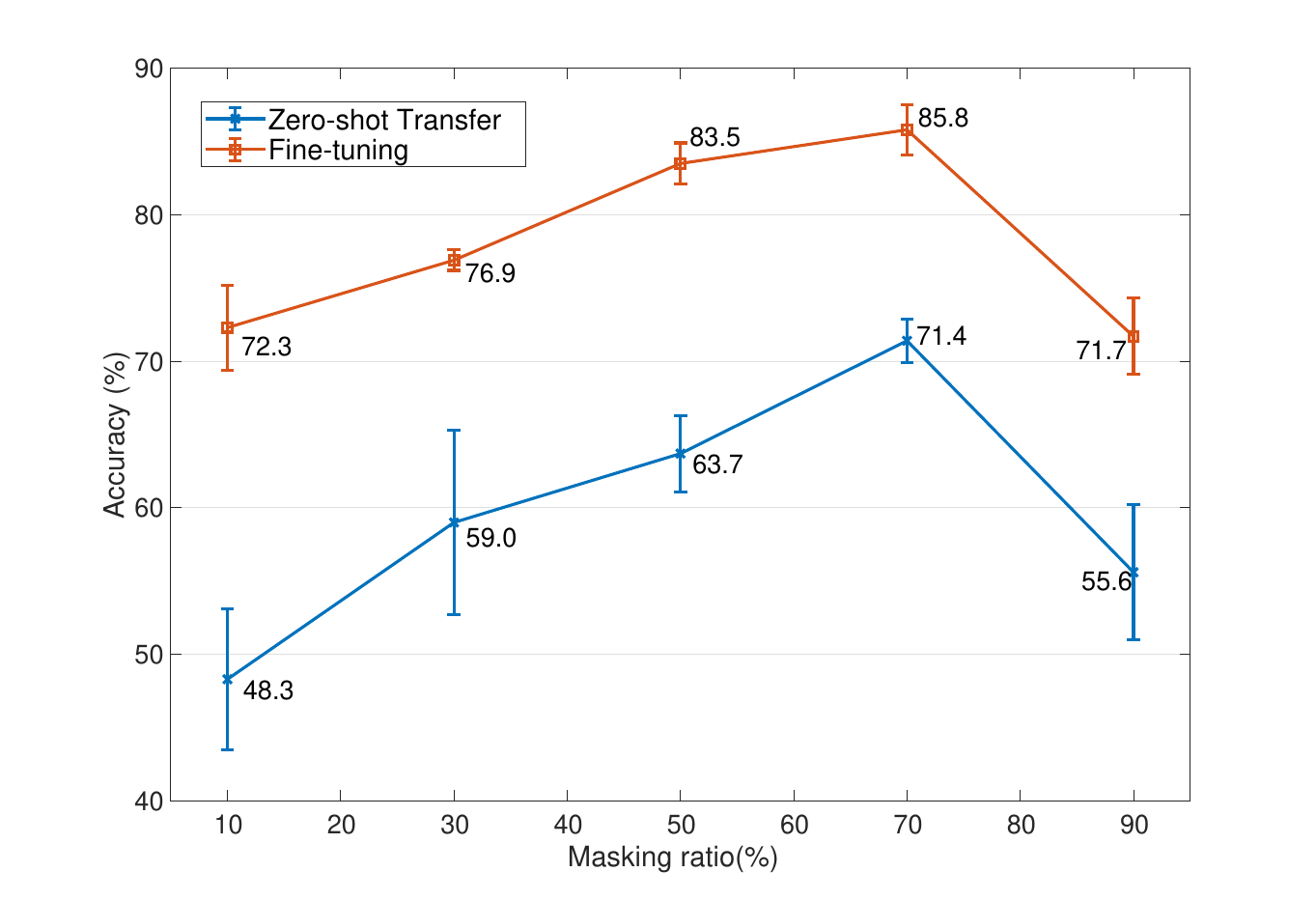}
	\caption{\textbf{\textit{Recognition results with different masking ratio}}. The x-axis represents the masking ratio ranging from 10\% to 90\%. The y-axis represents the recognition accuracy of incomplete tactile images from partial contact. The red and blue lines represent the results of fine-tuning and zero-shot transfer, respectively.}
	\label{fig:ratio}
\end{figure}

\begin{table}
	\centering

		\caption{Experimental results with different network structures }
		\label{tab:ablation}
        \scalebox{1}{
		\begin{tabular}{c| c | c }
			\hline
			\multirow{2}{*}{\makecell[c]{Network-\\structures}} & \multirow{2}{*}{{\makecell[c]{zero-shot transfer}}} & \multirow{2}{*}{{\makecell[c]{fine-tuning}}}
            \\&\multicolumn{1}{c|}{}
             \\
			\hhline{=|=|=}

		 \multirow{2}{*}{\makecell[c]{w/o classification head \\(MAE)}} & \multirow{2}{*}{\makecell[c]{31.9$\pm{0.4}$}} & \multirow{2}{*}{61.7$\pm{1.6}$}
            \\&\multicolumn{1}{c|}{} \\
		
		\hline

         \multirow{2}{*}{\makecell[c]{w/o reconstruction (ViT)}} & \multirow{2}{*}{\makecell[c]{56.2$\pm{2.1}$}} & \multirow{2}{*}{77.6$\pm{1.8}$}
            \\ &\multicolumn{1}{c|}{}\\
		\hline

  \multirow{2}{*}{\makecell[c]{TacMAE (ours)}} & \multirow{2}{*}{\makecell[c]{\textbf{71.4$\pm{\textbf{1.5}}$}}} & \multirow{2}{*}{\textbf{85.8$\pm{\textbf{1.7}}$}}
            \\&\multicolumn{1}{c|}{} \\
		
		\hline

		\end{tabular}}
\end{table}

\begin{figure*}[t]
	\centering
	\includegraphics[width=2\columnwidth]{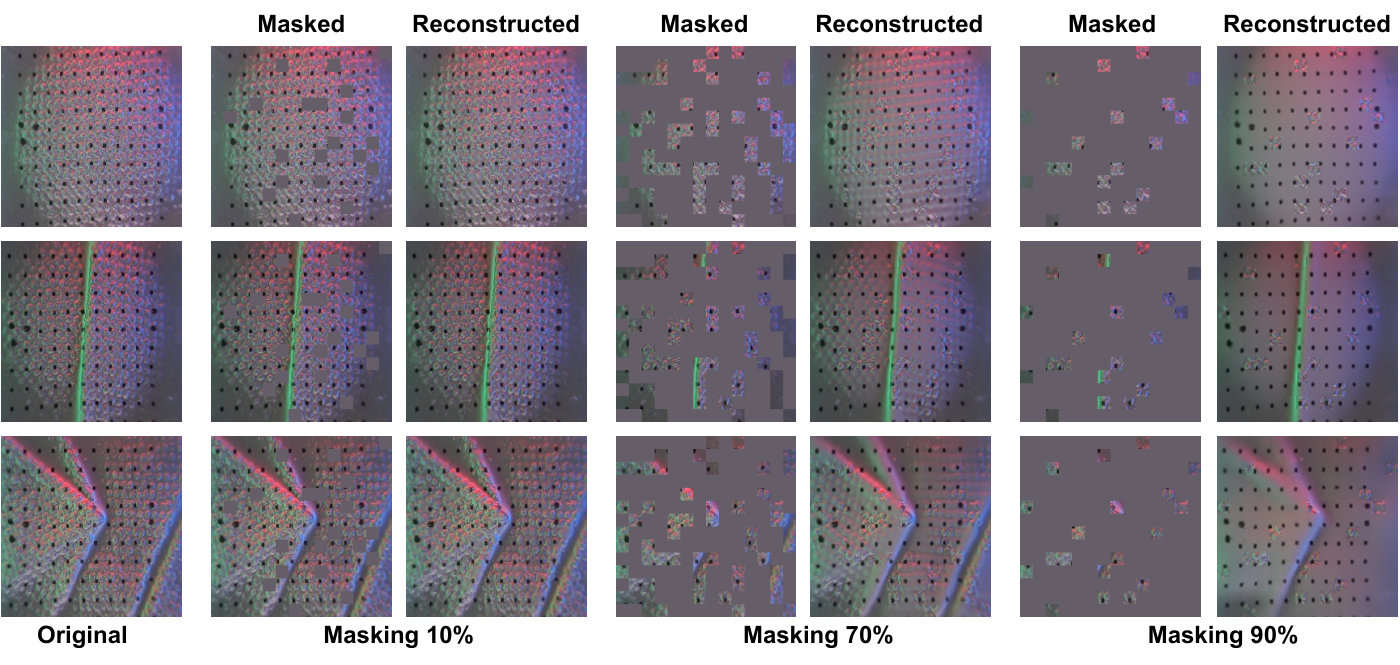}
	\caption{\textbf{Reconstruction of tactile images with different masking ratios.} From top to bottom: Tactile images collected when the fabric is flat, folded and laid randomly. Right six columns: the masked tactile images and reconstructed images with 10\%, 70\%, and 90\% masking ratio respectively.
	} 
	\label{fig:visulisation}
\end{figure*}

To validate the tactile representation capability of our proposed method, we test the results under two different settings, i.e., zero-shot transfer and fine-tuning in tactile recognition of incomplete tactile data.
The zero-shot learning usually involves the ability to recognise novel objects whose categories are not included in the training set. 
In this study, following~\cite{radford2021learning}, we extend the concept of zero-shot learning and investigate the generalisation to unobserved datasets. 
The motivation is to utilise it as a proxy to conduct unobserved tasks~\cite{larochelle2008zero}.
Concretely, the tactile representation is learnt by using the data from adequate contact, and the incomplete tactile data from partial contact are tested directly, without any additional training.
In addition to zero-shot learning, fine-tuning is a widely used method to evaluate the capability of representation. 
When compared to zero-shot learning, fine-tuning is a practical and adaptable approach that can modify the representation to suit the new dataset, thereby mitigating the failure in the representation learned during the pre-training phase.
Specifically, the representation model is trained by the data from adequate contact first. Then, a linear classifier on top of the learnt model is fine-tuned by using the incomplete data from partial contact.

In our framework, we use a high ratio of masking to simulate the contact area's absence for partial contact and reconstruct the masked patches to make the model learn the surface geometry and tactile features from the limited tactile signals.
First, we investigate how the masking ratio affects the results of tactile representation learning.
As shown in Fig~\ref{fig:ratio}, the recognition results are given for two different settings, zero-shot transfer and fine-tuning, with the masking ratio ranging from $10\%$ to $90\%$. 

When the masking ratio is $90\%$,  we see that recognition results are inferior to most others, with only 55.6\% in zero-shot transfer and 71.7\% in fine-tuning.
It indicates the challenge in establishing the correlation of surface geometry and predicting the missing patches with a small portion of unmasked tactile signals.
On the other hand, if the masking ratio is very small, (e.g., 10\%,) the reconstruction of the tactile signals is trivial as most of the tactile features are already present in the unmasked patches.
The optimal point for the masking ratio is around 70\%, where the recognition results are highest in our experiment, 71.4\% in zero-shot learning and 85.8\% in fine-tuning.
Moreover, the computation cost can also be reduced significantly because of the high masking ratio. 

\begin{table}
	\centering

		\caption{Recognition results compared with other baseline methods. }
		\label{tab:compare}
        \scalebox{1}{
		\begin{tabular}{c| c | c }
			\hline
			\multirow{2}{*}{\makecell[c]{Methods}} & \multirow{2}{*}{{\makecell[c]{zero-shot transfer}}} & \multirow{2}{*}{{\makecell[c]{fine-tuning}}}
            \\&\multicolumn{1}{c|}{}
             \\
			\hhline{=|=|=}

		 \multirow{2}{*}{\makecell[c]{Yuan \textit{et al.}~\cite{yuan2018active}}} & \multirow{2}{*}{\makecell[c]{26.4$\pm{2.8}$}} & \multirow{2}{*}{60.6$\pm{1.6}$}
            \\&\multicolumn{1}{c|}{} \\
		
		\hline

  \multirow{2}{*}{\makecell[c]{Cao \textit{et al.}~\cite{cao2020spatio}}} & \multirow{2}{*}{\makecell[c]{34.1$\pm{0.9}$}} & \multirow{2}{*}{74.5$\pm{1.8}$}
            \\ &\multicolumn{1}{c|}{}\\
		\hline

    \multirow{2}{*}{\makecell[c]{TacMAE (ours)}} & \multirow{2}{*}{\makecell[c]{\textbf{71.4$\pm{\textbf{1.5}}$}}} & \multirow{2}{*}{\textbf{85.8$\pm{\textbf{1.7}}$}}
            \\&\multicolumn{1}{c|}{} \\
		
		\hline

		\end{tabular}}
\end{table}

As shown in Fig~\ref{fig:visulisation}, we visualise the reconstructed tactile images with different masking ratios. 
When the masking ratio is at $10\%$, we are able to predict fine details of the missing patches.
However, if the masking ratio is increased to 70\%, the predicted details become less clear but the textures and geometry can be preserved.  
At a masking ratio of 90\%, only the outlines of the contact area can be reconstructed.
It also illustrates that a 70\% masking ratio could be a good compromise between the accuracy of the representation learnt and the masking ratio.

To further analyse how the proposed TacMAE method works in tactile representation learning, we conduct an ablation study. 
TacMAE consists of a reconstruction component and a classification component.
In the ablation study, we explore the effect of removing these two components, one at a time.
As shown in Table~\ref{tab:ablation}, our proposed TacMAE achieves the highest texture recognition results for  incomplete tactile data.
Concretely, the accuracy decreases by 15.2\% and 8.2\% in zero-shot transfer and fine-tuning respectively when we remove the reconstruction branch and use full patches for training. 
When the classification head is discarded, there is an obvious drop in performance, by 39.5\% and 24.1\%, respectively in zero-shot learning and fine-tuning.

\subsection{Comparison against other methods}
We compare TacMAE against the other existing methods in tactile texture recognition for the incomplete tactile data from partial contact.
Concretely, two CNN-based methods~\cite{yuan2018active,cao2020spatio} are tested in both zero-shot transfer and fine-tuning settings. 
In~\cite{yuan2018active}, a multi-label classification is performed and we modify the method for categorical classification.
In~\cite{cao2020spatio}, spatio-temporal attention is applied to address the salient features in a tactile sequence, and we remove the temporal attention function as the input in our experiment is a single tactile frame.

From Table~\ref{tab:compare}, we see that our proposed method achieves the highest recognition accuracy in both zero-shot learning and fine-tuning.
Specifically, we can notice that the baseline methods~\cite{yuan2018active,cao2020spatio} lack the ability to generalise in zero-shot learning, only obtaining 26.4\% and 34.1\% respectively. 
This is because the baseline methods, receiving the data for training from adequate contact, are unable to extract useful tactile features from incomplete tactile data due to differences in data distributions.
In our proposed framework, masking a portion of tactile signals and reconstructing the missing signals allow us to obtain semantic information of features from limited tactile signals. 
This enables TacMAE to possess strong robustness and generalisation ability.
Although the performance of baseline methods clearly improves after the fine-tuning where the recognition accuracies are 60.6\% and 74.5\%, respectively, it is still inferior to the performance of TacMAE.

\begin{figure}
	\centering
	\includegraphics[width=1\columnwidth]{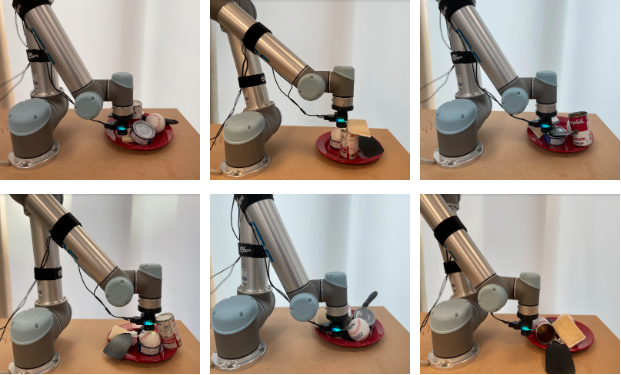}
	\caption{\textbf{\textit{Obstructed YCB objects.}} The objects are intentionally arranged with occlusion where the target object can only be partially reached by the tactile sensor. Top row: the robotic arm mounted with a GelSight sensor makes partial contact with the abrasive sponge, a tomato soup can and a tuna fish can, respectively. Bottom row: the GelSight sensor makes partial contact with a baseball, a spatula and a metal plate, respectively.  }
	\label{fig:YCB}
\end{figure}

\begin{table}[t]
	\centering
		\caption{The recognition results for YCB objects when partially contacted. }
		\label{tab:tactile_exploration}
        \scalebox{1}{
		\begin{tabular}{c| c|c|c  }
			\hline
			Object Category & Yuan \textit{et al.}~\cite{yuan2018active} & Cao \textit{et al.}~\cite{cao2020spatio} & TacMAE   \\
                \hhline{=|=|=|=}
            Abrasive sponge & 12/12 & 12/12 & 12/12 \\ \hline
            Tomato soup can & 4/12 & 5/12 & 12/12  \\ \hline
            Tuna fısh can & 9/12 & 10/12 & 10/12  \\ \hline
            Baseball & 11/12 & 12/12 & 12/12 \\ \hline
            Spatula & 3/12 & 4/12 & 9/12 \\ \hline
            Metal plate & 1/12 & 3/12 & 5/12 \\ \hline
            Average success rate & 55.6\% & 63.9\% & \textbf{83.3\%} \\
            \hline
		\end{tabular}}
\end{table}

\subsection{Transfer Learning by exploring YCB Objects}
The goal of this experiment is two-fold: 1) to demonstrate the cross-task knowledge transferability of our tactile representation method by performing an object recognition task with YCB objects~\cite{calli2015ycb}, and 2) to show that this method is able to improve the efficiency in active tactile exploration. 

\textbf{Experimental setup.} There are six YCB objects with different surface textures selected, including the abrasive sponge, tomato soup can, tuna fish can, baseball, spatula, and a metal plate.
Firstly, the tactile textures are collected from every single object to fine-tune the whole model.
Then, as shown in Fig~\ref{fig:YCB}, the objects are intentionally placed in such a way that the target object is obstructed by other objects.
As a result, in an active tactile exploration task, the target object can only be partially contacted by the tactile sensor without being relocated.
In the robotic experiment, a GelSight sensor is equipped on the UR5 robot arm as an end effector to obtain partial contacts for texture recognition.

\textbf{Experimental results.} In the robotic tactile exploration, we let the sensor to contact the objects partially to recognise the objects for 12 attempts per each object.
The contact location will be changed randomly by about 3mm on the horizontal plane in each attempt.
Particularly, the average contact area makes up about 11.5\% of the sensor's perception field in these contacts, while the other areas remain empty. 
Table~\ref{tab:tactile_exploration} compares the success rate of recognition against other baseline methods~\cite{yuan2018active,cao2020spatio}.
The results show that our TacMAE has the highest performance, 83.3\% recognition accuracy, in a total of 72 attempts. 
These results demonstrate the unique ability of TacMAE to transfer the learnt knowledge of tactile features to different kinds of objects.
It also indicates that TacMAE can improve the efficiency of active tactile exploration, i.e., obtaining more information with fewer attempts of touching, especially for the obstructed object without having to move it.

\section{Conclusion}
In this paper, we proposed TacMAE, a robust tactile representation method based on the MAE to effectively learn the features from partial contacts. 
During training, we create a simulation of partial contact by masking out a portion of tactile signals.
By reconstructing the missing signals via self-supervised learning, the model is capable to learn the surface geometry and the correlation between limited tactile cues.
The experimental results show that TacMAE obtains accurate tactile representations in both zero-shot learning and fine-tuning setting.
Furthermore, results on the YCB objects indicate the generalisation ability and knowledge transferability on different tasks.
Moreover, the ability to acquire knowledge from partial contact can increase the efficiency of tactile exploration, especially for obstructed objects.

TacMAE has the potential to be used as a  sensor-agnostic representation learning method by converting the signals into the format of images from any tactile sensor, not limited to the GelSight sensor.
We plan to explore these possibilities in our future work.
Moreover, we will investigate different downstream tasks using our proposed method, such as defect detection and robotic manipulation tasks.


{\small
	\bibliographystyle{ieeetr}
	\bibliography{reference.bib}
}

\end{document}